\documentclass[10pt,twocolumn,letterpaper]{article}

\usepackage{iccv}
\usepackage{times}
\usepackage{epsfig}
\usepackage{graphicx}
\usepackage{amsmath}
\usepackage{amssymb}
\usepackage{bbm}
\usepackage{adjustbox}
\usepackage{bbding}
\usepackage{multirow}
\usepackage[english]{babel}
\usepackage[utf8]{inputenc}
\usepackage{array}
\usepackage{multirow}
\usepackage{float}
\usepackage{color}
\usepackage{amsfonts,amssymb}
\usepackage{bm}
\usepackage{subfig}
\usepackage{algorithm,algorithmic}
\usepackage{lipsum}
\usepackage{pifont}
\newcommand{\cmark}{\ding{51}}%
\newcommand{\xmark}{\ding{55}}%
\usepackage{authblk}
\usepackage{setspace}

\usepackage{hyperref}
\hypersetup{breaklinks=true,letterpaper=true,colorlinks,bookmarks=false}
\iccvfinalcopy 

\def\iccvPaperID{2865} 

\ificcvfinal\fi

\begin{document}

\title{Morphable Detector for Object Detection on Demand}
\author[1]{Xiangyun Zhao}
\author[2]{Xu Zou}
\author[1]{Ying Wu} 
\affil[1]{Northwestern University}
\affil[2]{Huazhong University of Science and Technology}


\maketitle
\ificcvfinal\thispagestyle{empty}\fi

\begin{abstract}

Many emerging applications of intelligent robots need to explore and understand new environments, where it is desirable to detect objects of novel classes on the fly with minimum online efforts. This is an object detection on demand (ODOD) task. It is challenging, because it is impossible to annotate a large number of data on the fly, and the embedded systems are usually unable to perform back-propagation which is essential for training. Most existing few-shot detection methods are confronted here as they need extra training. We propose a novel morphable detector (MD), that simply ``morphs" some of its changeable parameters online estimated from the few samples, so as to detect novel classes without any extra training. The MD has two sets of parameters, one for the feature embedding and the other for class representation (called ``prototypes"). Each class is associated with a hidden prototype to be learned by integrating the visual and semantic embeddings. The learning of the MD is based on the alternate learning of the feature embedding and the prototypes in an EM-like approach which allows the recovery of an unknown prototype from a few samples of a novel class. 
Once an MD is learned, it is able to use a few samples of a novel class to directly compute its prototype to fulfill the online morphing process. We have shown the superiority of the MD in Pascal~\cite{everingham2010pascal}, COCO~\cite{lin2014microsoft} and FSOD~\cite{fan2020fsod} datasets. The code is available https://github.com/Zhaoxiangyun/Morphable-Detector.
	\vspace{-3mm}

\end{abstract}

\section{Introduction}

In applications, like robotics exploration and autonomous driving, the systems need to explore and understand new environments, where it is desirable to detect objects of novel classes on the fly with minimum online human supervision and interaction. This is an object detection on demand (ODOD) task. 
ODOD is very challenging because it is impossible to collect a large amount of data on the fly, and the computing resources are generally not powerful enough for computationally intensive and time-consuming training on board, not to mention that many embedded systems are unable to perform back-propagation which is essential for training. In embedded systems, the detection task is usually carried out on a computationally limited platform where the neural networks are locked after the system is built due to resource limits~\cite{alibaba2020mnn}. The prevailing few-shot detection (FSD)~\cite{kang2019few,yan2019meta,wang2019meta,wang2020frustratingly, xiao2020few} methods are confronted here as they generally need to perform extra training  
for objects from novel classes. 


\begin{figure}[t]
	\centering
	\includegraphics[scale=0.3]{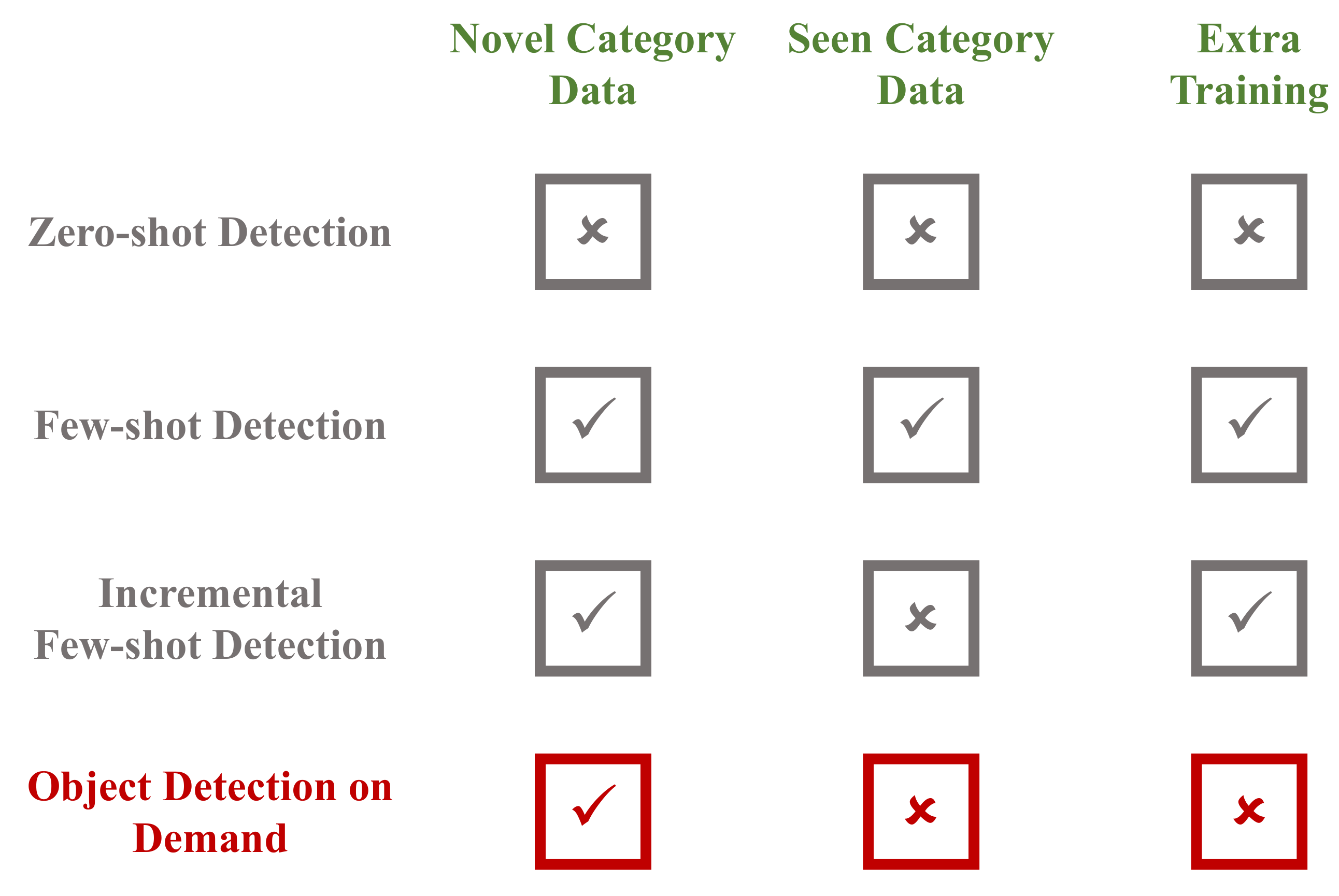}
	\caption{Comparison with other detection tasks. Different from other few-shot detection tasks, Object Detection on Demand requires no extra training.}
	\label{fig.task}
	\vspace{-3mm}
\end{figure}
To this end, 
we define Object Detection on Demand (ODOD) specifically as detecting the novel classes without extra training while preserving the existing knowledge, given (1) a detector offline trained using base class data, (2) no access to base class data (3) a few samples for novel classes. The ODOD can be regarded as a special few-shot detection task and the differences of ODOD from other detection tasks are listed in Fig~\ref{fig.task} .


The prevailing few-shot detectors (FSD) aim to train a detector using base class data and further train it with a few samples from novel classes. However, extra training is unfeasible in the ODOD task. Furthermore, to keep the performance for the base classes, these FSDs have to use the base class data in extra training, otherwise, they suffer from catastrophic forgetting~\cite{mccloskey1989catastrophic} - a significant performance degradation, when the
past data are not available. Other few-shot detectors~\cite{fan2020fsod, hsieh2019one} use a siamese network and take ``query-target" pairs as input so as to detect all instances of the ``target" object appearing in the ``query" image. However, as the target representation is always changing during the training, the model learns less discriminative representations. As a result, the model's generalizability to unseen samples of the base classes is limited. 

\begin{figure*}[t]
	\centering
	\includegraphics[scale=0.46]{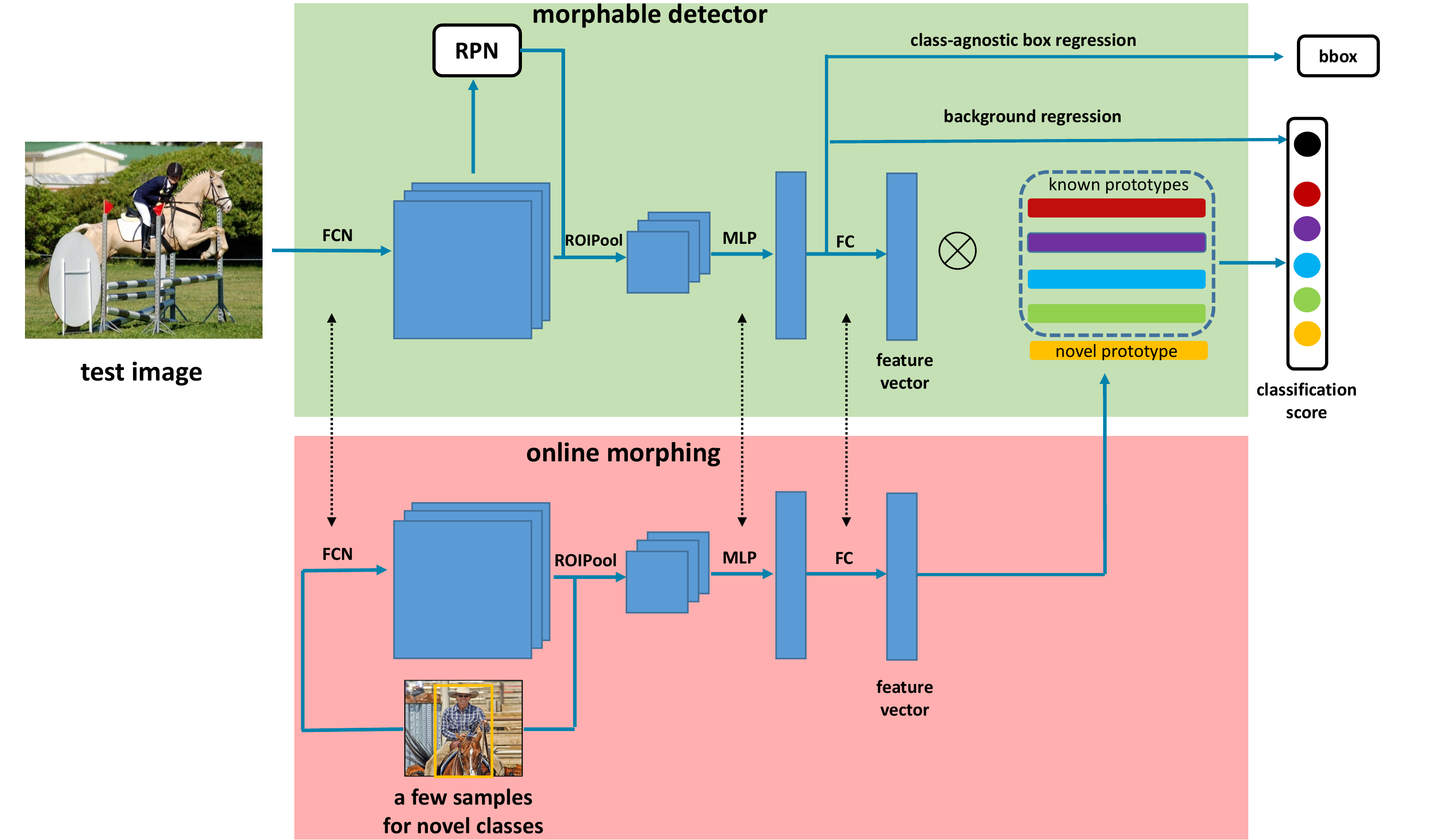}
	\caption{The proposed morphable detector (MD) structure. Given a trained MD, it is able to compute the representations (prototypes) for the novel classes using a few samples from novel classes (see sec.~\ref{sec.morphine}). Given a test image and proposals generated by RPN, the MD outputs the feature vectors, box regression, and background score for each proposal. The similarity between the feature vectors with the prototypes associated with each novel class is used to estimate the novel class posterior probability (see sec.~\ref{sec.morphine}).
	}
	\label{fig.framework}
	\vspace{-3mm}
\end{figure*}

In this paper, we present a novel morphable detector (MD) that simply ``morphs" some of its changeable parameters online estimated from the few samples, so as to detect novel classes without any extra training. Different from most existing object detectors, this novel MD has two sets of parameters, one for the feature embedding (i.e. the network parameters), and the other for class representation (called ``prototypes") as illustrated in Fig~\ref{fig.framework}. We view the MD for recognizing visual samples of different classes as if they live in a common space called feature space. Each class is associated with a prototype which is the targeted coordinate of each class in the feature space. Therefore, for each object proposal, the MD learns a feature vector whose similarity with prototypes is regarded as the foreground classification score. As it is hard to assign one prototype to the background, the MD directly regresses a background score from the visual features as shown in Fig~\ref{fig.framework}. Once an MD is learned, it is able to use a few samples of a novel class to directly compute its prototype to fulfill the online morphing process (details are in \ref{sec.morphine}).


 The learning of the MD is based on the alternate learning of the feature embedding and the prototypes in an EM-like approach as shown in Fig~\ref{fig.trianing}. The prototype is regarded as a hidden variable to be learned by integrating the visual and semantic embeddings. In "E"-step, we fix the network parameters and update the prototypes by combining the current prototypes and the feature vectors of the training samples on the trained model (details are in Sec.~\ref{sec.update}). In "M"-step, the prototypes are fixed and the network is trained using the updated prototypes. The prototypes are initialized with semantic vectors which bring useful external information from textual data. But note that directly using semantic vectors as prototypes without the proposed EM-like algorithm still suffers from limited generalizability (to novel classes) because the external information does not directly examine visual appearances while the model concerns itself with recognizing visual features. Therefore, the joint learning of the feature embedding and prototypes allows better recovery of an unknown prototype from a few samples of a novel class. Our approach is different from the existing approach, such as RepNet~\cite{karlinsky2019repmet} which learns representatives for each class from the visual data in an end-to-end training. The proposed MD learns the representatives (prototypes) by an EM-like approach where the visual and semantic information are integrated to improve the model's generalizability (to novel classes).

Overall, the contributions of this work are four-fold: 
\vspace{-2mm}
\begin{itemize}
    \item We study a special few-shot detection task, Object Detection on Demand, which is rarely discussed in the literature and can not be solved by many existing Few-shot detection methods. \vspace{-2mm}
    \item We present a novel morphable detector (MD) which can be online morphed to detect the novel classes without extra training. \vspace{-2mm}
    \item  We propose to learn the MD by joint visual and semantic embedding in an EM-like approach.  \vspace{-2mm}
    \item  Extensive experiments are performed on different datasets to demonstrate the superiority  of the  MD over other methods.\vspace{-2mm}
\end{itemize}

\section{Related Work}
\paragraph{Zero / few Shot Learning}
Zero Shot Learning (ZSL)~\cite{elhoseiny2017link,elhoseiny2013write,qiao2016less,reed2016learning,akata2015label,lampert2009learning,lampert2013attribute, zhao2019recognizing} has been widely studied for image recognition. It aims to 
recognize unseen classes without training samples. People usually leverage semantic information~\cite{elhoseiny2017link,elhoseiny2013write,qiao2016less,reed2016learning} or attributes representation~\cite{akata2015label,lampert2009learning,lampert2013attribute} for ZSL. Few Shot Learning aims to recognize a class with a few annotated samples. People try to address this problem by metric learning based approaches~\cite{vinyals2016matching,snell2017prototypical,oreshkin2018tadam,triantafillou2017few,hariharan2017low} or meta-learning based approaches~\cite{ravi2016optimization, zhang2018metagan, munkhdalai2017meta, munkhdalai2018rapid}. Different from them, in this work, we focus a more challenging object detection task. 
\vspace{-1mm}
\paragraph{Object Detection tasks} 
Most of existing detection methods~\cite{cai2016unified,cai2018cascade,lin2017feature,liu2016ssd,lin2017focal,redmon2016you,redmon2017yolo9000,redmon2018yolov3, zhao2020object, zhao2018pseudo, ren2015faster,girshick2014rich, girshick2015fast} focus on general object detection task where each category has large number of annotated data. However, when the labeled data are
scarce or not available, the models can overfit or fail to generalize. So people start to focus on zero-shot / few-shot detection~\cite{li2019zero, bansal2018zero, zhu2020don, kang2019few,wang2020frustratingly,wang2019meta,yan2019meta,xiao2020few, chen2018lstd} tasks where no example or a few examples for novel category are given. But the models can suffer from catastrophic forgetting~\cite{mccloskey1989catastrophic} when the
past data are not available. People start to work on incremental few-shot detection~\cite{perez2020incremental} task. Different from the above tasks, we focus on a more challenging task, Object Detection on Demand, where only a few samples for novel categories are given and no extra training is required. 
\vspace{-6mm}
\paragraph{Non-morphable Detector}
Recent works~\cite{zhu2020don, kang2019few,wang2020frustratingly,wang2019meta,yan2019meta,xiao2020few, chen2018lstd} have made significant progresses on few-shot detection tasks. They aim to leverage fully labeled seen category data to train a base model and adapt this model to novel classes using a meta feature
learner(i.e. extra training). However, extra training is unfeasible for emerging applications of robots. Different from them, we propose to train a morphable detector that can be online morphed to detect novel categories without any extra training. 
\vspace{-5mm}
\paragraph{Morphable Detector} One basic morphable detector is a zero-shot detector~\cite{li2019zero,zhu2019zero,bansal2018zero} which can detect the novel categories by leveraging semantic information without any annotated examples. However, as the semantic information does not directly examine the visual appearances, zero-shot detectors have limited generalization capability and the overall performance is far from satisfactory. Another morphable detector is some few-shot detector which takes ``query-target" pairs as input to detect all instances of the ``target" object appearing in the ``query" image.  But this model learns less discriminative representations as the target representations are always changing in training. RepNet~\cite{karlinsky2019repmet} propose to learn a few representatives for each category from the visual data. But learning from visual data is not enough for the model generalization. Different from them, we propose to leverage external semantic information~\cite{mikolov2013distributed} and present an EM-like approach to integrate the visual and semantic embeddings. 
\vspace{-2mm}

\section{Morphable Detector}
\begin{figure*}[t]
	\centering
	\includegraphics[scale=0.55]{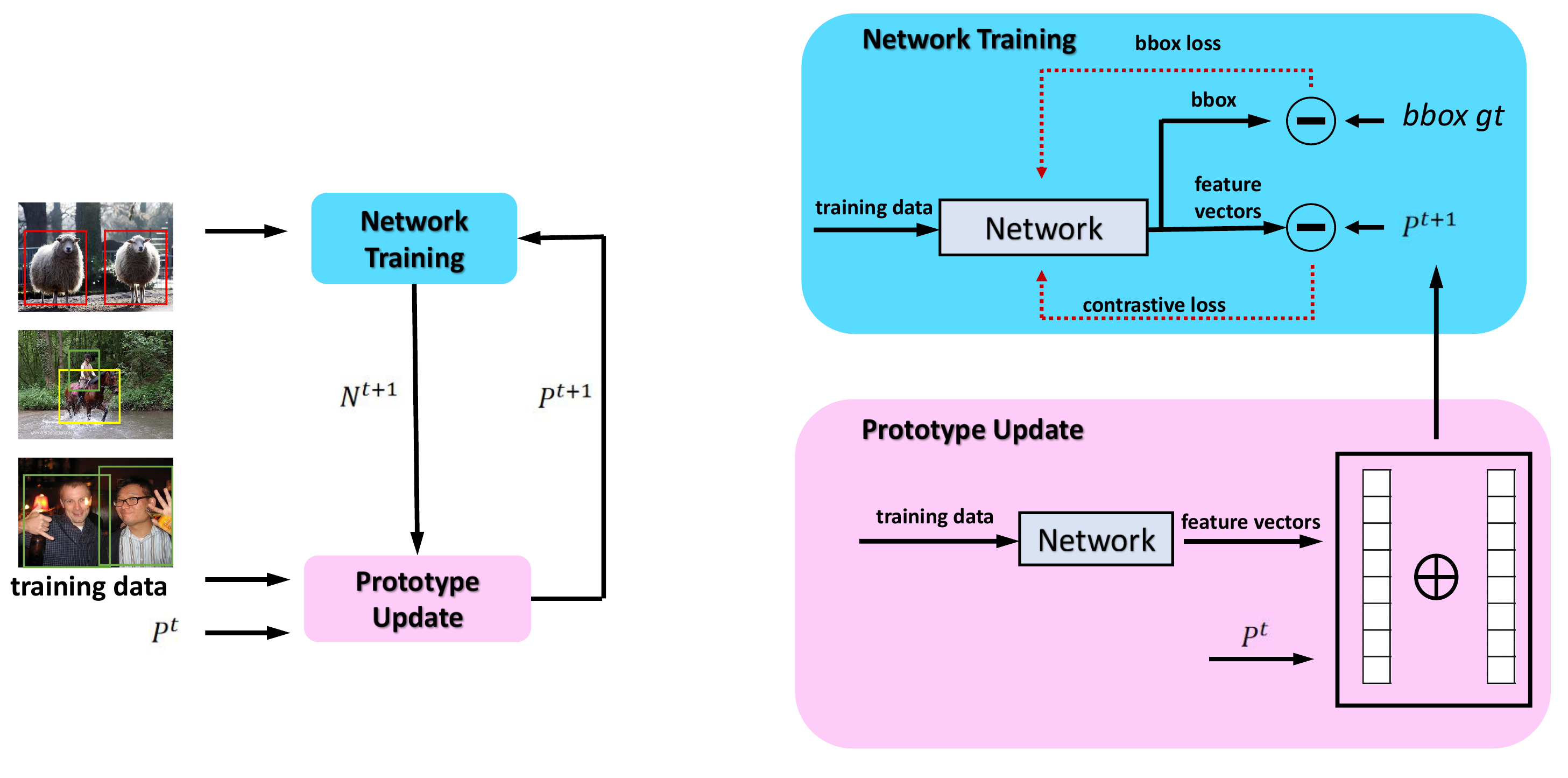}
	\caption{Training framework (EM-like approach). The learning of the MD is based on the alternate learning of the feature embedding and the prototypes in an EM-like approach. In ``E" step, with the network fixed, we compute the mean feature vector for each base class on the feature space to update the prototypes associated with that class (see sec. ~\ref{sec.update}). In ``M" step, with the prototypes fixed, we use the prototypes computed in ``E" step to train the network (see sec. ~\ref{sec.train}). }
	\label{fig.trianing}
	\vspace{-3mm}
\end{figure*}

\vspace{-2mm}
In this paper, we present a novel Morphable Detector (MD) which can be online morphed to detect the novel classes without extra training.  As illustrated in Fig~\ref{fig.trianing}, we propose to learn the prototypes and network parameters alternately in an EM-like approach, with the other fixed in each iteration. Fig~\ref{fig.framework} illustrates how the morphable detector(MD) is morphed to detect the novel classes given a few samples from novel classes. Once the MD is trained, the MD only needs to forward a few samples of a novel class through the network to compute its prototype (details are in sec.~\ref{sec.morphine}) to detect the novel classes.

\subsection{Basic Morphable Detector}

 We have base class set $\mathcal{C}_{base}$ and novel class set  $\mathcal{C}_{\rm novel}$, in which $\mathcal{C}_{\rm base} \cap \mathcal{C}_{novel} = \phi$. 
 We denote the base class dataset as $D_{\rm base}$ which consists of the training images and the corresponding box annotations. The MD framework applies to a variety of CNN-based detectors~\cite{ren2015faster, dai2016r, lin2017feature, cai2018cascade}. Here we instantiate the framework with Faster R-CNN(FRCNN)~\cite{ren2015faster} because it is a simple and widely used framework. The MD uses Region Proposal Network (RPN)~\cite{ren2015faster} to generate proposals and ROI pooling to extract the proposal features as illustrated in Fig.~\ref{fig.framework}. The MD has two sets of parameters: the network parameters and the class representation (called ``prototypes"). We denote the prototypes for base and novel classes as $\mathcal{P}_{\rm base}$ and $\mathcal{P}_{\rm novel}$ respectively. $\mathcal{P}_{\rm base}$ is learned by joint visual and semantic embeddings in the EM-like approach. Once the MD is trained,  $\mathcal{P}_{\rm novel}$ can be computed by forwarding the samples from novel classes through the trained network.  Specifically, a trained MD's parameters consist of the network parameter $\Theta$ and prototypes $\mathcal{P}_{\rm base}$. Once $\mathcal{P}_{\rm novel}$ is computed, the MD can be online morphed to a new detector whose parameters consist of $\Theta$, $\mathcal{P}_{\rm base}$ and  $\mathcal{P}_{\rm novel}$. As a result, the new detector can detect novel classes.

 \subsection{Training of Morphable Detector}
In the Morphable Detector (MD), each class is associated with one prototype. Suppose we have prototypes for base classes $\mathcal{P_{\rm base}} = \{p_{j}\}$ where $j$ indicates the class. MD generates the proposals $\{x_i,y_i\} \in {\rm ROI}$ by the RPN~\cite{ren2015faster}. The MD learns prototypes for the base classes and a feature space where the feature vector of a given sample is expected to be close to the corresponding prototype while far away from prototypes of other classes. The objective is to maximize the likelihood:
 \begin{equation}
\hspace{-4mm}\sum_{x_i, y_i \in {\rm ROI}, y_i >0}P(y_i|p_i)P(p_i|x_i) + \hspace{-6mm} \sum_{x_i, y_i \in {\rm ROI}, y_i = 0}P(y_i|x_i)
\label{eq.likelihood}
\end{equation}

where $P(p_i|x_i)$ or $P(y_i|x_i)$ is determined by the network output and $P(y_i|p_i)$ is determined by the prototype associated with each class.

To maximize the above likelihood, we regard the prototype as a hidden variable and propose to learn it by integrating the visual and semantic embedding. The feature embedding and the prototypes are alternately learned in an EM-like approach where the prototypes are initialized with semantic vectors in the initial training and recursively updated over iterations. 
 In ``E"-step, the network parameters $\Theta$ are fixed so the $P(p_{i}|x_i)$ is a constant in Eq~\ref{eq.likelihood}. It aims to update the prototypes for next iterative training. So, the ``E" step takes provided base class data $D_{\rm base}$, the trained model $\mathcal{N}^t$ and the current prototypes $\mathcal{P}_{base}^t$ as inputs, then outputs updated prototypes $\mathcal{P}^{t+1}_{\rm base}$.
   \begin{equation}
\mathcal{P}^{t+1}_{\rm base}= E(D_{\rm base},  \mathcal{N}^t,  \mathcal{P}_{base}^t).
\label{eq.mapping1}
\end{equation}

In ``M" step, with the prototypes fixed, $P(y_i|p_i)$ is a constant. The optimization is essentially equivalent to the maximum-likelihood estimation of the network parameters. Therefore, the "M" step takes the training data $D_{\rm base}$ and the prototypes $\mathcal{P}_{\rm base}^{t+1}$ for base classes as input, and outputs a newly trained network model $\mathcal{N}^{t+1}$.
 
  \begin{equation}
\mathcal{N}^{t+1}= M(D_{\rm base}, \mathcal{P}_{\rm base}^{t+1}).
\label{eq.mapping1}
\end{equation}

 The model usually converges after several iterations.

 \vspace{-3mm}
 
 \subsubsection{Network Training}
 
 \label{sec.train}
 Given the extracted proposals $\{x_i,y_i\} \in {\rm ROI}$,  the deep visual features for each proposal $x_i$ are extracted as $\phi(x_i)$. As it is hard to assign a prototype to the background, the MD directly regresses a background score from the visual features. $\phi(x_i)$ is forwarded through two separate fully connected layers to obtain the background score $b_i \in  \mathbb{R}^1$ and the feature vector $f_i  \in  \mathbb{R}^d$, where $d$ is the dimension of the feature vector. The network is trained with a prototype-based contrastive loss which consists of two terms: the foreground loss and background loss. The foreground loss encourages the feature vector of the proposal $\{x_i, y_i \in {\rm ROI} \}$ to be close to the corresponding prototype while far away from other prototypes if the proposal belongs to the foreground (i.e. $y_i >0$). So, the foreground loss is defined as


\begin{small}
\begin{equation}
L_{FG} = \sum_{y_i >0}^{}-{\rm log}(\frac{\exp(f_i \cdot p_{y_i})}{\exp(b_i) + \sum_{p_m\in \mathcal{P}_{\rm base}}\exp( f_i \cdot p_{m}) }),
\label{eq.loss}
\end{equation}
\end{small} where $p_{y_i}$ is the prototype corresponding to class $y_i$. When the proposal belongs to the background, the contrastive loss encourages the background score to be high. The background loss is defined as

\begin{small}
\begin{equation}
L_{BG} = \sum_{y_i =0}^{}-{\rm log}(\frac{\exp{(b_i)}}{\exp(b_i) + \sum_{p_m\in \mathcal{P}_{\rm base}}\exp( f_i \cdot p_{m}) }).
\label{eq.loss}
\end{equation}
\end{small}

The overall loss is sum of the foreground loss, background loss, and a class-agnostic bounding box regression loss~\cite{ren2015faster}
 \begin{equation}
L = L_{\rm BG} + L_{\rm FG} + L_{\rm bbox}.
\label{eq.cl}
\end{equation}

 \subsubsection{Prototype update}
 \label{sec.update}
We denote one ground truth box for class $j$ as $x_i \in g_j$ and the feature vectors of it on the network $\mathcal{N}$ is $\mathcal{N}(x_i)$. Then we compute the mean feature vector $v_j$ of all samples from each class $j$ as \begin{equation}
  v_{j} = \frac{1}{|g_{j}|}\sum_{x_i \in \rm g_{j}}\mathcal{N}^t(x_i).
\end{equation}
Then we use the mean feature vector $v_{j}$ to compute the new prototype $p^{t+1}_{j}$ by fusing it with the current prototypes $p^t_{j}$ (associated with class $j$) by weighted element-wise sum,
\begin{equation}
  p_{j}^{t+1} = (1-\lambda)v_{j} + \lambda p_{j}^t,
\label{eq.update}
\end{equation}
where $\lambda$ is a constant between 0 and 1. Note that before the element-wise sum, both prototypes and mean feature vectors are normalized.

\subsection{Online Morphing}
\label{sec.morphine}
The online morphing is to compute the new prototypes for the novel classes as shown in Fig~\ref{fig.framework}. Suppose we have a ground truth box $x_i \in g_{j}$ from novel class $j$, and forward the samples through the network to get the feature vectors $\mathcal{N}(x_i)$. The mean feature vector of all samples belonging to a novel class $j$ is used as the new prototype for that class, 
 \begin{equation}
  p_{j} = \frac{1}{|g_{j}|}\sum_{x_i \in \rm g_{j}}\mathcal{N}(x_i) .
\end{equation}

Now, we have the $\mathcal{P}_{\rm novel} = \{p_j\}$ where $j$ is the class index for a novel class, so the novel class can be detected. Given a test image, the RPN first generates the proposals $x_i \in {\rm ROI}$, and get the bbox score $b_i$ and feature vector $f_i$. The class posterior probability for class $j$ is,
\begin{small}
\begin{equation}
\frac{\exp(f_i \cdot p_{j})}{\exp(b_i) + \sum_{p_m\in \mathcal{P}_{\rm base} \cup \mathcal{P}_{\rm novel}}\exp( f_i \cdot p_{m}) }.
\label{eq.loss}
\end{equation}
\end{small}
Then, the detected boxes are obtained by setting a threshold for the class score like other detectors~\cite{ren2015faster}.


\begin{table}[t]
\centering
{
	\center
		\begin{tabular*}{0.49\textwidth}{@{\extracolsep{\fill}}p{2cm}<{\centering}|p{1.6cm}<{\centering}|p{1.5cm}<{\centering}|p{1.5cm}<{\centering}}
\hline
\multicolumn{1}{l|}{{Method}}  & {\footnotesize Extra training} & $\rm AP$ & $\rm AP_{0.5}$    \\
\hline
\hline

\multicolumn{1}{c|}{FRCNN~\cite{ren2015faster}}             &     \cmark      & -          & 23.0  \\
\multicolumn{1}{c|}{LSTD~\cite{chen2018lstd}}             &    \cmark      & -          & 24.2  \\
\multicolumn{1}{c|}{FSOD method~\cite{fan2020fsod}}               &   \xmark     & -          & 27.5          \\
\hline
\multicolumn{1}{c|}{Visual (ImageNet)} &   \xmark      & 10.1         & 16.3 \\
\multicolumn{1}{c|}{Visual (FRCNN)} &  \xmark      & 15.5         & 22.6 \\
\multicolumn{1}{c|}{MD (concat)} &   \xmark     & 17.3 &29.9 \\
\multicolumn{1}{c|}{MD ($\lambda=0$)} &    \xmark      & 21.5 & 36.2   \\
\multicolumn{1}{c|}{MD ($\lambda=0.3$)} &  \xmark      & 21.3 & 35.9  \\
\multicolumn{1}{c|}{MD ($\lambda=0.7$)} &   \xmark      & 21.6 & 36.3  \\
\multicolumn{1}{c|}{MD (iter1)} & \xmark      & 18.2 & 31.2  \\
\multicolumn{1}{c|}{MD (iter2)} &   \xmark       & 21.9 & 36.7 \\
\multicolumn{1}{c|}{MD} &    \xmark      & \textbf{22.2} & \textbf{37.1} \\

\hline
\end{tabular*}
}
\caption{Comparison of our method with different baselines and different variants on FSOD Dataset}
\label{tab:fsod}
\end{table}

\begin{table}[t]
\centering
{
	\center
		\begin{tabular*}{0.49\textwidth}{@{\extracolsep{\fill}}p{0.7cm}<{\centering}|p{0.7cm}<{\centering}p{0.7cm}<{\centering}|p{0.7cm}<{\centering}p{0.7cm}<{\centering}|p{0.7cm}<{\centering}}
\hline
\multicolumn{1}{c|}{\multirow{2}{*}{Method}}  & \multicolumn{2}{c|}{Split 1} & \multicolumn{2}{c|}{Split 2} & Ave  \\
                  & $\rm AP$       & $\rm AP_{0.5}$  & $\rm AP$ & $\rm AP_{0.5}$ & $\rm AP_{0.5}$\\

\hline
\hline
\multicolumn{1}{c|}{OSOD~\cite{hsieh2019one}}                  & -          & -  & -          & -& 22.0
\\
\multicolumn{1}{c|}{MD (iter1)}                  & 20.2& 32.9 &21.1 &32.6& 32.8
\\
\multicolumn{1}{c|}{MD}                  & 
\textbf{21.5}&
\textbf{33.0}&
\textbf{24.9}&
\textbf{36.1}& \textbf{34.6}\\

\hline
\end{tabular*}
}
\caption{One-shot detection performance comparison on the first two splits of COCO dataset for the novel classes.}
\label{tab:coco1}
\end{table}
\begin{table*}[t]
\centering
{
	\center
		\begin{tabular*}{\textwidth}{@{\extracolsep{\fill}}p{1.5cm}<{\centering}|p{1.5cm}<{\centering}|p{1.5cm}<{\centering}|p{1.5cm}<{\centering}|p{1.7cm}<{\centering}|p{1.5cm}<{\centering}|p{1.7cm}<{\centering}|p{1.7cm}<{\centering}|p{2cm}<{\centering}}
\hline
\multicolumn{1}{c|}{\multirow{2}{*}{\footnotesize  Method}}   & {\footnotesize  Ours (1-shot)}  &{\footnotesize FSView~\cite{xiao2020few} (1-shot)}& {\footnotesize LSTD~\cite{chen2018lstd} (10-shot)} & {\footnotesize MetaYOLO~\cite{kang2019few} (10-shot)}& {\footnotesize MetaDet~\cite{wang2019meta} (10-shot)}& { \footnotesize MetaRCNN~\cite{yan2019meta}  (10-shot)} & {\footnotesize TFA w/fc~\cite{wang2020frustratingly} (10-shot)}&{\footnotesize TFA-w/cos~\cite{wang2020frustratingly} (10-shot)}   \\
\hline		
\hline

\multicolumn{1}{c|}{$\rm AP$}                  & \underline{9.7}          & 4.5 &3.2 &5.6&7.1&8.7& \textbf{10.0}&\textbf{10.0}\\
\multicolumn{1}{c|}{$\rm AP_{0.5}$}                    & \underline{15.0}          & 12.4&8.1&12.3&14.6 &\textbf{19.1}&-&- \\
\multicolumn{1}{c|}{$\rm AP_{0.75}$}                         & \textbf{9.9}          & 2.2&2.1 &4.6&6.1 &6.6&9.2& \underline{9.3}       \\

\hline
\end{tabular*}
}
	\vspace{-2mm}
\caption{Performance comparison with state-of-the-arts on COCO dataset for novel classes in split 3. The best one and the second-best one are highlighted in \textbf{bold} and \underline{underlined} respectively.}
\label{tab:coco2}
\end{table*}
	\vspace{-2mm}
\begin{table*}[t]
\centering
{
	\center
		\begin{tabular*}{\textwidth}{@{\extracolsep{\fill}}p{1.5cm}<{\centering}|p{1.5cm}<{\centering}|p{1.5cm}<{\centering}|p{1.5cm}<{\centering}|p{1.7cm}<{\centering}|p{1.5cm}<{\centering}|p{1.7cm}<{\centering}|p{1.7cm}<{\centering}|p{2cm}<{\centering}}
\hline
\multicolumn{1}{c|}{{Method}}  & {\footnotesize Ours}  &{\footnotesize FSView~\cite{xiao2020few}}&{\footnotesize LSTD~\cite{chen2018lstd}} & {\footnotesize MetaYOLO~\cite{kang2019few}}& {\footnotesize MetaDet~\cite{wang2019meta}}&{\footnotesize MetaRCNN~\cite{yan2019meta}}& {\footnotesize TFA w/fc~\cite{wang2020frustratingly}}&{\footnotesize TFA w/cos~\cite{wang2020frustratingly}}   \\
\hline
\hline

\multicolumn{1}{c|}{Split 1}                  &    \textbf{53.2}    & 24.2 &8.4 &14.8 & 18.9&19.9&22.9&25.3\\
\multicolumn{1}{c|}{Split 2}                    &      \textbf{41.6}      & 21.6  & 11.4 &15.7 & 21.8 &10.4 & 16.9 & 18.3\\
\multicolumn{1}{c|}{Split 3}                         & \textbf{38.6}       & 21.2 &12.6  &21.3 & 20.6 &14.3&15.7&17.9      \\

\hline
\end{tabular*}
}
\caption{Performance comparison with state-of-the-arts PASCAL VOC dataset for novel classes in three splits.}
\vspace{-0.2cm}
\label{tab:voc}
\end{table*}



\section{Experiments}


To evaluate our morphable detector (MD) on Object Detection on Demand (ODOD), we begin with an evaluation on a challenging large-scale dataset FSOD~\cite{fan2020fsod} which benchmarks the performance of detectors on few-shot detection setting. Then we evaluate on two widely used datasets MS COCO~\cite{lin2014microsoft} and Pascal VOC datasets~\cite{everingham2010pascal}, and compare against state-of-the-arts on few-shot detection setting. Finally, as a by-product, we compare against state-of-the-arts on zero-shot detection setting. 

\subsection{Experiments on FSOD dataset}

\paragraph{Dataset and implementation} Few Shot Object Detection (FSOD)~\cite{fan2020fsod} dataset is proposed to evaluate the detector which is trained using base class data and evaluated on the novel classes. This dataset contains 1000 classes
with 800/200 split for training and test set respectively. There is no overlap between the training and test classes. There are 52350 images with 147489 annotated boxes in the training set and 14152 images with 35102 annotated boxes in the test set. We use the Region Proposal Network and class-agnostic regression from Faster-RCNN in our model. Following the training strategies in ~\cite{fan2020fsod}, we use ResNet-50 as our backbone and pretrain the model on the COCO dataset. Then we train the model using the base class data which contains 800 classes and test on the test set which contains 200 novel classes. We randomly select 5 samples for each novel class as known samples for the novel classes as ~\cite{fan2020fsod}. We train the model with batch size 4 for 50k iterations with a learning rate 0.002 and another 20k iterations with a learning rate 0.0002. All models are evaluated using standard ${\rm AP_{0.5}}$ and ${\rm AP}$. The dimension size of the feature embedding and semantic vectors~\footnote{We use the extracted semantic vectors from \url{https://github.com/agnusmaximus/Word2Bits}} is 200.

\vspace{-3mm}
\paragraph{Comparison with baselines}
We first compare against different baselines which use different prototype initializations. Then, we evaluate the MD model after each iteration. 
\begin{itemize}
    \item \textbf{Visual (ImageNet/FRCNN)}. In our MD, the prototypes are initialized using semantic vectors. In the experiments, we compare against MD variants using visual features for prototype initialization. To get the visual features for each class, we forward the base class ground truth samples through a trained Faster-RCNN model or ImageNet pre-trained model to get the visual features whose dimension is 1024 in our experiment. Then we use the mean feature vectors for each class as the prototype for that class to train the MD. 
    \item \textbf{Iterative results}. We evaluate the MD model's performance after each iteration. 
\end{itemize}

Table~\ref{tab:fsod} summarizes the comparisons against the baselines. Directly using ImageNet~\cite{deng2009imagenet} or FRCNN~\cite{ren2015faster} features as prototypes does not work well. The results show that the model learned only using visual features can not be well generalized to novel classes. The semantic information learned from text data provides useful information about the relationship between different classes. So, the MD (iter 1) which uses semantic vectors as prototypes obtains significant improvements over the model only using visual features. This verifies that semantic vectors can help improve the model's generalizability to novel classes. Note that the semantic information does not examine the visual appearances. As a result, the overall performance is still limited. To overcome this limitation, our MD is learned by integrating the visual and semantic embeddings. The results show that the model's performance can be significantly improved by the joint visual and semantic embedding. We empirically find that the performance can not be further improved after 2-3 iterations so we set the number of iterations to be 3.
\vspace{-3mm}
\paragraph{Ablation study}

To study the best way to combine the mean feature vector and the prototypes to compute the new prototypes, we compare ``element-wise sum" and ``concatenate". After the initial training, we concatenate or element-wise sum the mean feature vectors of the ground truth samples and the prototypes (i.e., the first item and second item in Eq.~\ref{eq.update}). The experimental results show that the element-wise sum is a better combination way, so the MD uses element-wise sum in the remaining experiments. Then, we compare the MDs using different $\lambda$ in Eq.~\ref{eq.update}, and we find that $\lambda=0.5$ works the best.         

\begin{figure}[t]
	\centering
	\includegraphics[scale=0.55]{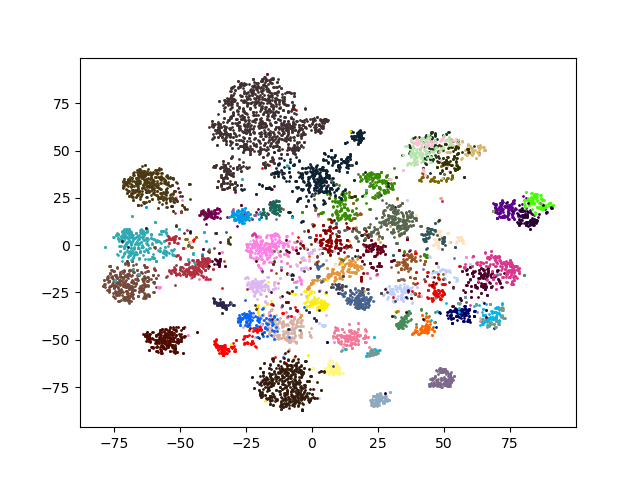}
	\caption{ t-SNE visualization of feature embeddings of objects
from randomly selected 50 novel classes in FSOD test set on the learned MD.}
	\label{fig.embedding}
\end{figure}

\paragraph{Comparison with state-of-the-arts}
We compare against three state-of-the-arts: FSOD method~\cite{fan2020fsod}, FRCNN~\cite{ren2015faster}, LSTD~\cite{chen2018lstd}. FRCNN and LSTD results are reimplemented by ~\cite{fan2020fsod}. Our MD obtains $\rm AP_{0.5}$ 10 points improvements over FSOD and 13 points improvement over LSTD. Unlike FSOD method~\cite{fan2020fsod} which takes 'query-target' pairs as input to train the model, our morphable detector learns more discriminative features and leverages useful external textual information. Therefore, our detector has better generalizability to novel classes. To verify that the MD can be well generalized to novel classes, we randomly selected 50 classes from the test set and visualize the feature vectors by t-SNE tool~\cite{van2008visualizing}. Fig~\ref{fig.embedding} shows that objects from most novel classes are clustered together on the learned feature space.

\vspace{-3mm}
\paragraph{Computation time} It takes around 0.04
seconds in ResNet 50 and 0.09 seconds in ResNet 101 to
add a new class using a single RTX 3090 GPU.
\subsection{Experiments on Pascal and COCO}
\paragraph{Datasets and implementation}On Pascal VOC~\cite{Everingham15}, VOC 07 and 12 train/val sets are used for training and VOC 2007 test set is used for testing. In order to fairly compare with the state-of-the-art methods~\cite{kang2019few,chen2018lstd,yan2019meta}, we follow \cite{kang2019few} to use the same novel splits. 
On MS COCO~\cite{lin2014microsoft}, we follow~\cite{lin2017feature} and train the model using the union of 80k train
images and a 35k subset of val images (trainval35k~\cite{bell2016inside}), and report the testing performed on a 5k subset of val images
(minival). We group the 80 classes in COCO dataset into 5 different semantic clusters and randomly select two classes from each cluster as novel classes(i.e. 10 classes in total). We randomly select two splits using this strategy. In the split 1, we select ``\textit{bicycle}", ``\textit{car}", ``\textit{dog}", ``\textit{sheep}",  ``\textit{frisbee}", ``\textit{surfboard}", ``\textit{pizza}", ``\textit{laptop}",  ``\textit{microwave}" and ``\textit{refrigerator}" as the novel classes. In the split 2, we select ``\textit{car}", ``\textit{train}", ``\textit{boat}", `` \textit{dog}", `` \textit{horse}", ``\textit{skateboard}", ``\textit{sandwich}", ``\textit{pizza}",  ``\textit{keyboard}" and ``\textit{microwave}" as the novel classes. To fairly compare with most few-shot detectors~\cite{kang2019few, chen2018lstd,yan2019meta,wang2019meta,wang2020frustratingly, xiao2020few}, we also perform experiments using the same split as them (i.e., using 20 Pascal VOC classes as novel classes). Following these few-shot detectors, we ignore the novel classes annotations in training and randomly select one example as a given example for each novel class in testing. So, we perform one-shot detection experiments. We use ResNet-101~\cite{he2016deep} as the backbone and train the model with batch size 4 for 50k iterations and 160k iterations for Pascal VOC and COCO separately. The initial learning rate is 0.002 and it is decreased to 0.0002 after 40k and 120k iterations for Pascal and COCO. We use the same embedding size and semantic vectors used in FSOD experiments.


 \begin{figure*}[t]
	\centering
	\includegraphics[scale=0.55]{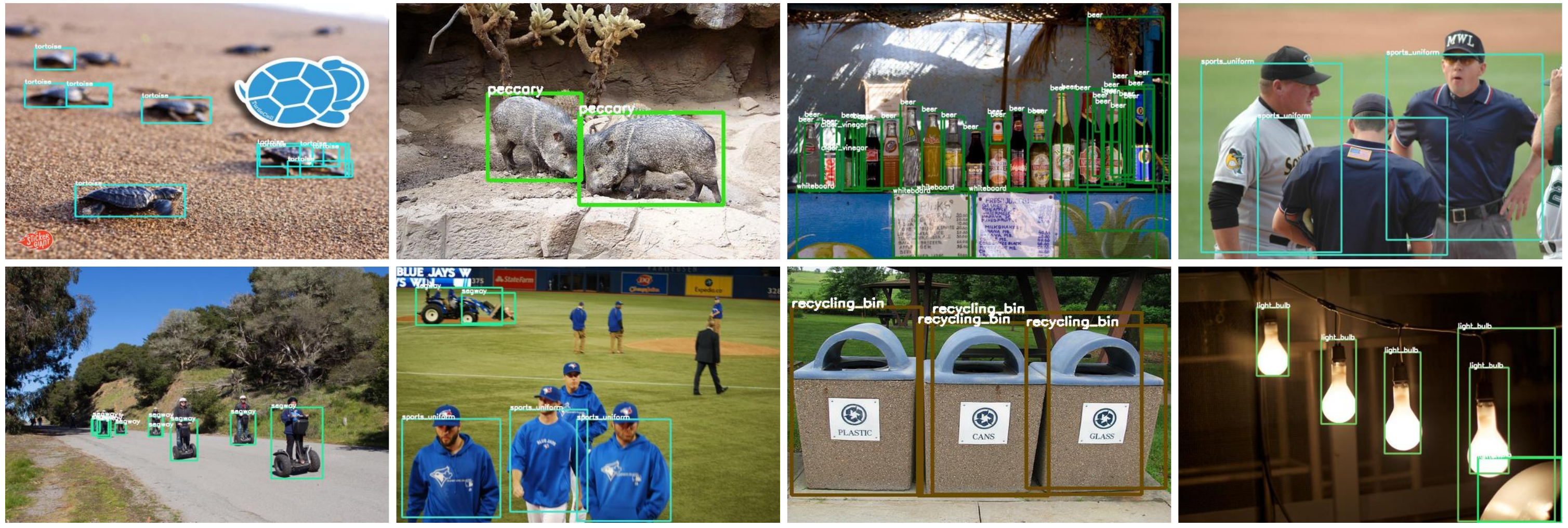}
	\caption{Some qualitative results of our proposed Morphable Detector on FSOD test set.}
	\label{fig.example}
\end{figure*}
\subsection{Comparison on novel classes}
Table~\ref{tab:coco1} summarizes the comparisons on COCO dataset for split 1 and 2. Over iterations, the MD's performance consistently improves on the two splits. This verifies the effectiveness of the proposed EM-like approach. We also use the average performance to compare against one-shot-detector~\cite{hsieh2019one} which perform their experiments on another four random splits. Same with ~\cite{fan2020fsod}, OSOD~\cite{hsieh2019one} takes "query-target" pairs as input. So, our MD has better generalizability to the novel classes than~\cite{hsieh2019one}. Table~\ref{tab:coco2} summarizes the comparison against several state-of-the-arts~\cite{kang2019few, chen2018lstd, yan2019meta,wang2019meta,wang2020frustratingly, xiao2020few} on COCO dataset for split 3. Among them, only FSView~\cite{xiao2020few} reported their performance on one-shot detection settings. For others, they reported their 10-shot performance in their paper. Our method outperforms FSView by a large margin in the 1-shot detection setting and outperforms most of the others even though we only use 1-shot data. More importantly, as all these methods need extra training, they can not be deployed on the embedded systems as our method. Note that the performance may vary for different splits. The reason is that as the prototypes are initialized with semantic vectors, the relationship between base and
novel classes can influence the MD’s performance. In
the first two splits, the base and novel classes are split based
on semantic clusters of the classes, so the results of them
can be obviously better than those of split 3. Table~\ref{tab:voc}  summarizes the comparison against the state-of-the-arts on three splits of the Pascal VOC dataset. Our MD outperforms state-of-the-arts by a large margin on the three splits. Our MD leverages semantic and visual information to help generalize the trained model to novel classes. 

\subsection{Comparison on base classes}
Table~\ref{tab:coco_seen} and~\ref{tab:voc_seen} summarize the comparison against our baselines and state-of-the-arts on Pascal and COCO datasets for base classes. Our MD obtains obviously a bit better performance on the base classes over iterations. This verifies the proposed EM-like algorithm can help improve the model's generalizability to unseen samples of base classes. 
Our MD performs much better than OSOD~\cite{hsieh2019one} which takes ``query-target" pairs as input. This verifies that OSOD~\cite{hsieh2019one} learns much less discriminative features for base classes. Compared with state-of-the-arts few-shot detectors, our MD performs the best for the base classes. The reason is that these FSD models take a small number of base class data to further train the model so these models can easily overfit to the small data. Compared with FRCNN~\cite{ren2015faster}, our model can still outperform it. This shows the advantage of our MD over FRCNN on the base classes. Note that FRCNN can not be generalized to novel classes without extra training.  


\begin{table}[t]
\centering
\footnotesize
{
	\center
		\begin{tabular*}{0.49\textwidth}{@{\extracolsep{\fill}}p{0.3cm}<{\centering}|p{0.3cm}<{\centering}p{0.6cm}<{\centering}|p{0.3cm}<{\centering}p{0.6cm}<{\centering}|p{0.3cm}<{\centering}p{0.6cm}<{\centering}|p{0.7cm}<{\centering}}
\hline
\multicolumn{1}{c|}{\multirow{2}{*}{Method}}  & \multicolumn{2}{c|}{Split 1} & \multicolumn{2}{c|}{Split 2} & \multicolumn{2}{c|}{Split 3} & Ave  \\
                  & $\rm AP$       & $\rm AP_{0.5}$  & $\rm AP$ & $\rm AP_{0.5}$ & $\rm AP$ & $\rm AP_{0.5}$& $\rm AP_{0.5}$\\

\hline
\hline

\multicolumn{1}{c|}{FRCNN~\cite{ren2015faster}}                  & 

37.3&
59.9&
36.9&
58.7&
37.0&
59.1 & 59.2
\\
\multicolumn{1}{c|}{OSOD~\cite{hsieh2019one}}                  & -          & -  & -          & -& - & - & 40.9
\\
\multicolumn{1}{c|}{MD (iter1)}                  & 37.5&
60.6
&\textbf{37.2}
&\textbf{59.3}
&37.2
&59.0 & 59.6
\\
\multicolumn{1}{c|}{MD}                  & 

\textbf{37.8}&
\textbf{60.7}&
36.9&
59.0&
\textbf{37.5}&
\textbf{59.2} & \textbf{59.7}
\\
\hline
\end{tabular*}
}
	\vspace{-2mm}
\caption{Performance comparison on the COCO dataset for base classes.}
\label{tab:coco_seen}
\end{table}
\begin{table}[t]
\centering
{
	\center
		\begin{tabular*}{0.49\textwidth}{@{\extracolsep{\fill}}p{1cm}<{\centering}|p{1cm}<{\centering}|p{1cm}<{\centering}|p{1cm}<{\centering}|p{1cm}<{\centering}}
\hline
\multicolumn{1}{c|}{{Method}}  & {Split 1} & Split 2 & Split 3 & Ave   \\
\hline
\hline

\multicolumn{1}{c|}{MetaRCNN~\cite{yan2019meta}}            
&64.8& - & - &64.8\\
\multicolumn{1}{c|}{TFA w/cos~\cite{wang2020frustratingly}}            
&79.2& - & - &79.1\\
\multicolumn{1}{c|}{OSOD~\cite{hsieh2019one}}                  & -          & -  & -& 60.1\\

\multicolumn{1}{c|}{MD (iter1)}            
&80.2&81.6&
78.4&80.1 \\
\multicolumn{1}{c|}{MD}            
&\textbf{80.7}&\textbf{82.1} &\textbf{79.2}&\textbf{80.7}\\
\hline
\end{tabular*}
}
\caption{Performance comparison on the PASCAL VOC dataset for base classes. }
\label{tab:voc_seen}
\vspace{-5mm}

\end{table}

\begin{table}[t]
\centering
{
	\center
		\begin{tabular*}{0.44\textwidth}{@{\extracolsep{\fill}}p{3cm}<{\centering}|p{2cm}<{\centering}|p{2cm}<{\centering}}
\hline
\multicolumn{1}{c|}{{Method}}  & Recall@100 & $\rm AP_{0.5}$   \\
\hline
\hline
\multicolumn{1}{c|}{SB~\cite{bansal2018zero}}            
&22.4& 0.7 \\
\multicolumn{1}{c|}{DSES~\cite{bansal2018zero}}            
&27.2& 0.54 \\
\multicolumn{1}{c|}{TD~\cite{li2019zero}}            
&34.3 & - \\
\multicolumn{1}{c|}{DELO~\cite{zhu2020don}}            
&33.5& 7.6\\

\multicolumn{1}{c|}{MD (split 1)}            
&44.8& 9.4 \\
\multicolumn{1}{c|}{MD (split 2)}            
&\textbf{47.2}& \textbf{9.8} \\
\multicolumn{1}{c|}{MD (split 3)}            
&27.0& 4.2 \\
\hline
\end{tabular*}
}
\caption{Comparison under zero shot detection setting on COCO dataset for novel classes. }
\label{tab:zero}
\vspace{-2mm}

\end{table}
\subsection{Comparison with zero shot detectors}
As a by-product, we also perform the MD under zero-shot detection setting. In training, we use the semantic vectors for the base classes as prototypes to train the MD. In testing, we use the semantic vectors for the novel classes as novel prototypes. Table~\ref{tab:zero} summarizes the comparison against state-of-the-art zero shot detectors using standard evaluation metrics recall@100 and $\rm AP_{0.5}$. Our MD obtains very impressive results on the first two splits. The performance on the split 3 is not as good as the other two splits because the relationship between base and novel classes on split 3 does not help much as the other splits. Our MD obtains an average of 39.7 Recall@100 which is better than other zero-shot detectors.



\section{Conclusion}
In this paper, we focus on a very challenging task: object detection on demand (ODOD) task. The prevailing FSD methods can not solve this problem well as ODOD requires no extra training. We propose a novel morphable detector (MD), that simply ``morphs" some of its changeable parameters online estimated from the few samples, so as to detect novel categories without any extra training. The learning of the MD is based on the alternative learning of the feature embedding and the prototypes in an EM-like approach which allows better recovery of an unknown prototype from a few samples of a novel category. Extensive experiments are performed to demonstrate the superiority of the MD.

\paragraph{Acknowledgements} We thank Pengbo Zhao for the suggestions on the writing. This work was supported in part by National Science Foundation grant IIS-1619078, IIS-1815561, and IIS-2007613. 

\clearpage

{\small
\bibliographystyle{ieee_fullname}
\bibliography{egbib}
}

\end{document}


\title{Supplementary Material - Morphable Detector for Object Detection on Demand}

\author{First Author\\
Institution1\\
Institution1 address\\
{\tt\small firstauthor@i1.org}
\and
Second Author\\
Institution2\\
First line of institution2 address\\
{\tt\small secondauthor@i2.org}
}

\maketitle
\ificcvfinal\thispagestyle{empty}\fi
\section{Split details}
 \begin{figure*}
	\centering
	\includegraphics[scale=0.7]{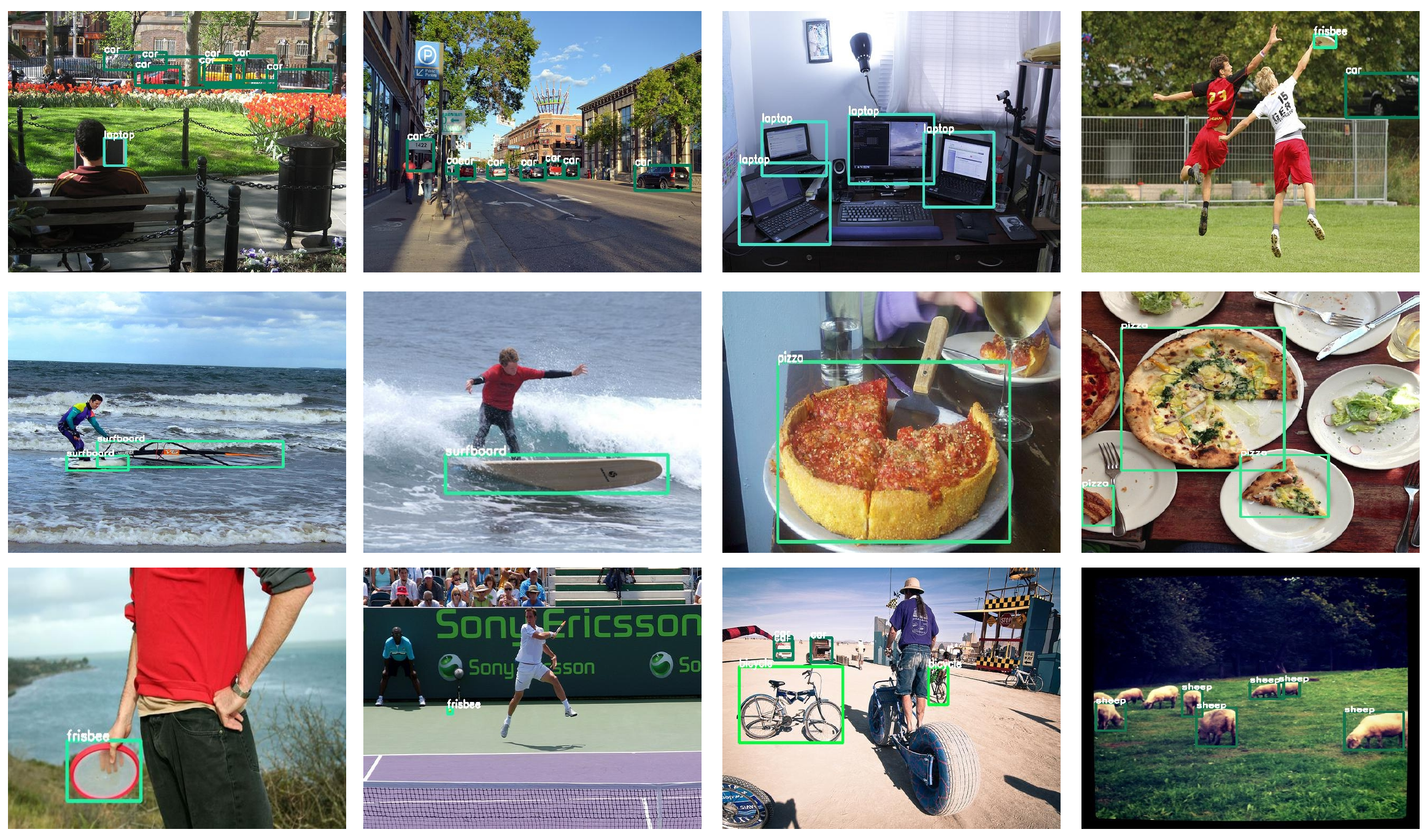}
	\caption{Qualitative results of our proposed Morphable Detector on COCO 2017 validation set. The model is trained in split 1 and we only visualize the qualitative results for the novel categories}
	\label{fig.coco_visual}
	\vspace{-3mm}
\end{figure*}
On the VOC dataset, we follow most few shot detectors~\cite{yan2019meta,kang2019few,wang2020frustratingly} and use the same splits as them. The results details are listed in Table~\ref{tab:voc}. On the COCO dataset, we group the 80 categories into 5 different clusters and randomly select two categories from each cluster as the novel categories. In the split 1, we select \textit{bicycle} \textit{car} \textit{dog} \textit{sheep} \textit{apple} \textit{frisbee} \textit{surfboard} \textit{pizza} \textit{laptop} \textit{microwave} \textit{refrigerator} as the novel categories. The results details are in Table~\ref{tab:coco1}.  In the split 2, we select car \textit{train} \textit{boat} \textit{dog} \textit{horse} \textit{skateboard} \textit{sandwich} \textit{pizza} \textit{keyboard} \textit{microwave} as the novel categories. The results details are in Table~\ref{tab:coco2}. In the split 3, we use all PASCAL VOC categories as the novel categories. The results details are in Table~\ref{tab:coco3}.
\section{Qualitative Results}
\begin{figure*}[t]
	\centering
	\includegraphics[scale=0.7]{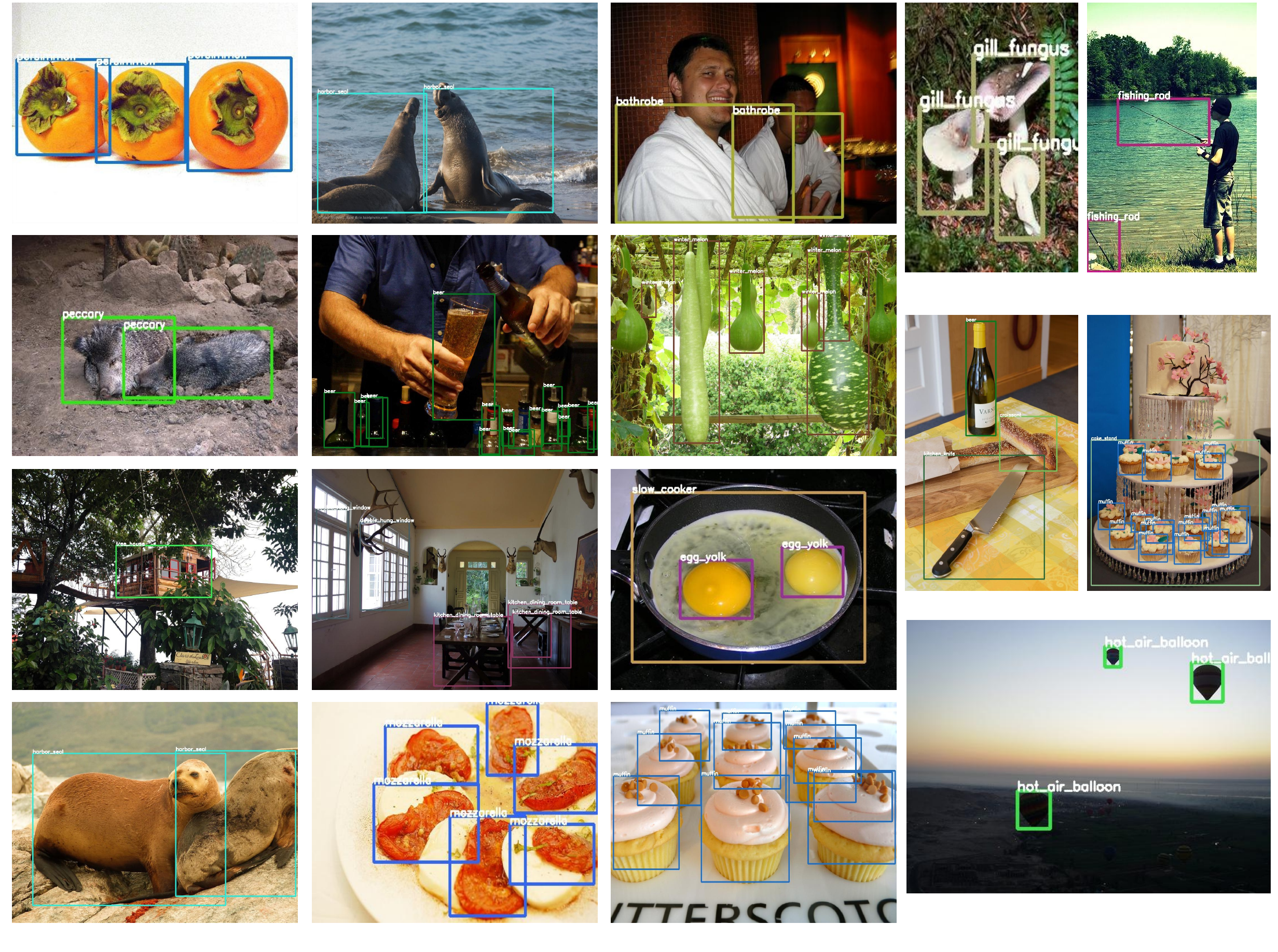}
	\caption{Qualitative results of our proposed Morphable Detector on FSOD test set.}
	\label{fig.fsod}
	\vspace{-3mm}
\end{figure*}

More qualitative results on FSOD test set~\cite{fan2020fsod} are in Figure~\ref{fig.fsod}. Qualitative results on COCO validation set are in Figure~\ref{fig.coco_visual}. The model is trained in split 1 and we only visualize the qualitative results for the novel categories. 

\begin{table*}[t]
\centering
\scriptsize
\setlength{\tabcolsep}{0.7mm}
{
	\center
		\begin{tabular*}{\textwidth}{@{\extracolsep{\fill}}p{0.7cm}<{\centering}|p{0.7cm}<{\centering}p{0.7cm}<{\centering}p{0.7cm}<{\centering}p{0.7cm}<{\centering}p{0.7cm}<{\centering}p{0.7cm}<{\centering}|p{0.7cm}<{\centering}p{0.7cm}<{\centering}p{0.7cm}<{\centering}p{0.7cm}<{\centering}p{0.7cm}<{\centering}p{0.7cm}<{\centering}|p{0.7cm}<{\centering}p{0.7cm}<{\centering}p{0.7cm}<{\centering}p{0.7cm}<{\centering}p{0.7cm}<{\centering}p{0.7cm}<{\centering}}
\hline
\multicolumn{1}{l|}{\multirow{2}{*}{Method}}  & \multicolumn{6}{c|}{Split 1} & \multicolumn{6}{c|}{Split 2} & \multicolumn{6}{c}{Split 3}  \\
                  &bird & bus & cow& mbike& sofa& mean& aero& bottle& cow &horse &sofa &mean& boat& cat& mbike& sheep& sofa &mean\\

\hline
\hline
\multicolumn{1}{l|}{Meta-YOLO~\cite{kang2019few}}                  &13.5 & 10.6 &30.5 &13.8& 4.3& 14.8 &11.8 &9.1 &15.6& 23.7& 18.2& 15.7& 10.8& 44.0& 17.8& 18.1& 5.3& 19.2 \\
\multicolumn{1}{l|}{Meta-RCNN~\cite{yan2019meta}}                  &6.10 & 32.8 &15.0 &35.4 &0.2 &19.9 &23.9& 0.8& 23.6& 3.1& 0.7& 10.4& 0.6& 31.1& 28.9& 11.0& 0.1& 14.3 \\
\multicolumn{1}{l|}{MD}                  &10.9
&42.0&24.9 &55.0 &5.4 &27.6 &62.2& 58.7& 58.6&65.8&20.5& 53.2& 
25.9
&71.3
&59.9
&60.0
&13.5
&46.1
\\
\hline
\end{tabular*}
}
\caption{The results on the PASCAL VOC dataset for novel categories. }
\label{tab:voc}
\end{table*}

\begin{table*}[t]
\centering
\setlength{\tabcolsep}{1.2mm}
{
	\center
		\begin{tabular*}{\textwidth}{@{\extracolsep{\fill}}p{1cm}<{\centering}|p{0.7cm}<{\centering}p{0.7cm}<{\centering}p{0.7cm}<{\centering}p{0.7cm}<{\centering}p{0.7cm}<{\centering}p{0.7cm}<{\centering}p{1.2cm}<{\centering}p{0.7cm}<{\centering}p{0.7cm}<{\centering}p{1.2cm}<{\centering}p{1.2cm}<{\centering}p{0.5cm}<{\centering}}
\hline
               
 & {\footnotesize bicycle}  &{\footnotesize car}&{\footnotesize dog} & {\footnotesize sheep}&{\footnotesize apple}& {\footnotesize frisbee}&{\footnotesize surfboard}& {\footnotesize pizza}&{\footnotesize laptop}&{\footnotesize microwave} &{\footnotesize refrigerator}  &{\footnotesize ave}   \\
\hline
\hline
$\rm AP$ & 9.70
&15.5
&26.1
&24.5
&3.1
&14.0
&17.3
&18.3
&27.1
&17.0
&31.0
&18.5 \\
$\rm AP_{50}$&
22.7
&33.1
&41.1
&39.2
&5.4
&31.6
&24.6
&29.0
&41.9
&22.8
&45.1
&30.6 \\
$\rm AP_{75}$&
6.60
&12.4
&28.3
&26.2
&3.1
&10.8
&24.2
&20.0
&30.5
&20.0
&37.8
&20.0 \\
\hline
\end{tabular*}
}
\caption{The results in the split 1 on the COCO dataset for the novel categories.}
\vspace{-0.2cm}
\label{tab:coco1}
\end{table*}

\begin{table*}[t]
\centering
\setlength{\tabcolsep}{0.7mm}
{
	\center
		\begin{tabular*}{\textwidth}{@{\extracolsep{\fill}}p{1cm}<{\centering}|p{0.7cm}<{\centering}p{0.7cm}<{\centering}p{0.7cm}<{\centering}p{0.7cm}<{\centering}p{0.7cm}<{\centering}p{1.2cm}<{\centering}p{1.2cm}<{\centering}p{0.7cm}<{\centering}p{1.2cm}<{\centering}p{1.2cm}<{\centering}p{0.7cm}<{\centering}}
\hline
                               
 & {\footnotesize car}  &{\footnotesize train}&{\footnotesize boat} & {\footnotesize dog}&{\footnotesize horse}& {\footnotesize skateboard}&{\footnotesize sandwich}& {\footnotesize pizza}&{\footnotesize keyboard}&{\footnotesize microwave} &{\footnotesize ave}   \\
\hline
\hline
$\rm AP$ &
19.2
&19.5
&31.0
&23.4
&17.3
&10.5
&3.35
&21.8
&10.1
&7.62
&16.4
\\
$\rm AP_{50}$&
38.4
&29.8
&48.9
&39.9
&30.6
&18.0
&6.18
&29.1
&16.4
&12.3
&27.0 \\
$\rm AP_{75}$&
17.0
&22.0
&34.1
&24.1
&18.3
&10.8
&2.93
&28.2
&10.1
&9.16
&17.7 \\
\hline
\end{tabular*}
}
\caption{The results in the split 2 on the COCO dataset for the novel categories.}
\vspace{-0.2cm}
\label{tab:coco2}
\end{table*}
\begin{table*}[t]
\centering
\scriptsize
\setlength{\tabcolsep}{1.2mm}
{
	\center
		\begin{tabular*}{\textwidth}{@{\extracolsep{\fill}}p{0.7cm}<{\centering}|p{0.7cm}<{\centering}p{0.7cm}<{\centering}p{0.5cm}<{\centering}p{1.0cm}<{\centering}p{1.0cm}<{\centering}p{0.4cm}<{\centering}p{0.5cm}<{\centering}p{0.5cm}<{\centering}p{0.5cm}<{\centering}p{0.4cm}<{\centering}p{0.5cm}<{\centering}p{0.6cm}<{\centering}p{0.6cm}<{\centering}p{0.4cm}<{\centering}p{0.5cm}<{\centering}p{0.5cm}<{\centering}p{0.5cm}<{\centering}p{0.5cm}<{\centering}p{0.5cm}<{\centering}p{0.3cm}<{\centering}p{0.5cm}<{\centering}}
\hline

 & {\footnotesize person}  &{\footnotesize bicycle}&{\footnotesize car} & {\footnotesize motorbike}&{\footnotesize airplane}& {\footnotesize bus}&{\footnotesize train}& {\footnotesize boat}&{\footnotesize bird}&{\footnotesize cat} &{\footnotesize dog}&{\footnotesize horse}&{\footnotesize sheep} & {\footnotesize cow}&{\footnotesize bottle}& {\footnotesize chair}&{\footnotesize sofa}& {\footnotesize plant}&{\footnotesize table}&{\footnotesize tv} &{\footnotesize ave}    \\
\hline
\hline
$\rm AP$ &
13.0
&4.2
&18.5
&3.7
&11.8
&8.7
&8.9
&3.1
&1.7
&24.7
&9.25
&14.3
&7.85
&12.5
&8.4
&3.4
&10.7
&3.4
&3.4
&23.2
&9.7
\\
$\rm AP_{50}$
&20.1
&9.5
&33.9
&7.9
&16.6
&12.5
&13.8
&7.01
&2.71
&39.6
&13.8
&21.9
&11.6
&18.9
&13.0
&7.33
&19.5
&6.45
&5.87
&34.2
&15.8
\\
$\rm AP_{75}$
&14.9
&2.7
&17.7
&3.5
&14.1
&9.9
&9.5
&2.4
&1.7
&28.4
&10.1
&16.5
&8.8
&13.9
&9.2
&2.7
&9.8
&3.4
&3.1
&24.5
&10.3
\\
\hline
\end{tabular*}
}
\caption{Performance details in the split 3 on the COCO dataset for the novel categories.}
\vspace{-0.2cm}
\label{tab:coco3}
\end{table*}
{\small
\bibliographystyle{ieee_fullname}
\bibliography{egbib}
}